\documentclass{article}

\usepackage[preprint]{neurips_2026}

\usepackage[utf8]{inputenc} 
\usepackage[T1]{fontenc}    
\usepackage{hyperref}       
\usepackage{url}            
\usepackage{booktabs}       
\usepackage{amsfonts}       
\usepackage{nicefrac}       
\usepackage{microtype}      
\usepackage{xcolor}         
\usepackage{amsmath,amssymb}
\usepackage{graphicx}
\usepackage{pifont}         
\usepackage[capitalize]{cleveref}
\newcommand{\cmark}{\ding{51}}
\newcommand{\xmark}{\ding{55}}
\usepackage{wrapfig}
\usepackage{bbm}

\title{Real2Sim in HOI: Toward Physically Plausible HOI Reconstruction from Monocular Videos}

\author{
Yubo Zhao$^{1,2}$ \quad Yujin Chai$^2$ \quad Yunao Dong$^1$ \quad \textbf{Chengfeng Zhao}$^1$\\
\textbf{Zijiao Zeng}$^2$ \quad \textbf{Yuan Liu}$^1$ \quad \textbf{Chi-Keung Tang}$^1$\\
$^1$The Hong Kong University of Science and Technology \quad $^2$Tencent IEG \\
\texttt{yzhaodx@connect.ust.hk}
}

\begin{document}

\maketitle

\begin{figure}[h]
    \vspace{-20pt}
    \centering
    \includegraphics[width=\linewidth]{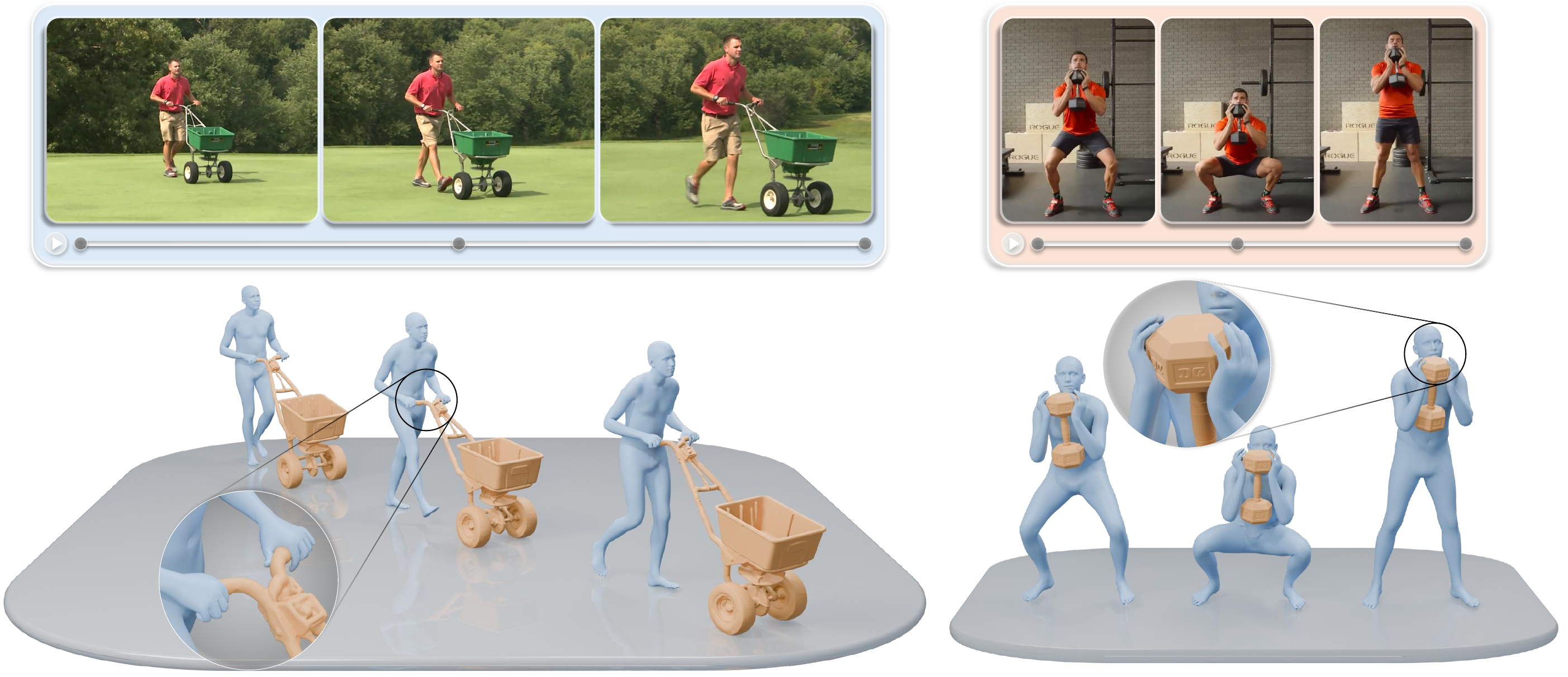}
    \caption{\textbf{Reconstructing interaction, not just trajectories.}
    From monocular RGB videos, \textit{HA-HOI} reconstructs simulation-ready 4D human-object interaction.
}
    \label{fig:teaser}
\end{figure}

\begin{abstract}

Recovering 4D human-object interaction (HOI) from monocular video is a key step toward scalable 3D content creation, embodied AI, and simulation-based learning. Recent methods can reconstruct temporally coherent human and object trajectories, but these trajectories often remain visual artifacts while failing to preserve stable contact, functional manipulation, or physical plausibility when used as reference motions for humanoid-object simulation. This reveals a fundamental interaction gap: HOI reconstruction should not stop at tracking a human and an object, but should recover the relation that makes their motion a coherent interaction.
We introduce \textit{HA-HOI}, a framework for reconstructing physically plausible 4D HOI animation from in-the-wild monocular videos. Instead of treating the human and object as independent entities in an ambiguous monocular 3D space, we propose a \emph{human-first, object-follow} formulation. The human motion is recovered as the interaction anchor, and the object is reconstructed, aligned, and refined relative to the human action. The resulting kinematic trajectory is then projected into a physics-based humanoid-object simulation, where it acts as a teacher trajectory for stable physical rollout.
Across benchmark and in-the-wild videos, \textit{HA-HOI} improves human-object alignment, contact consistency, temporal stability, and simulation readiness over prior monocular HOI reconstruction methods. By moving beyond visually plausible trajectory recovery toward physically grounded interaction animation, our work takes a step toward turning general monocular HOI videos into scalable demonstrations for humanoid-object behavior.

Project Page: \href{https://knoxzhao.github.io/real2sim_in_HOI/}{\nolinkurl{https://knoxzhao.github.io/real2sim_in_HOI/}}
\end{abstract}

\section{Introduction}

4D human-object interaction (HOI) reconstruction aims to recover temporally coherent and spatially consistent motions of humans and objects from visual observations. Converting videos into animatable human-object motion, it provides a key foundation for 3D content creation, AR/VR, embodied AI, and simulation-based learning. In particular, reinforcement learning has greatly benefited from kinematic human motion data as reference demonstrations for acquiring human-like behaviors~\citep{peng2018deepmimic,peng2021amp,xu2025intermimic}. However, obtaining reliable 4D HOI animation remains difficult. Traditional pipelines typically rely on human and object motion capture, multi-view RGB-D systems, or extensive manual post-processing. While effective in controlled environments, such setups are expensive, labor-intensive, and difficult to scale.

Monocular RGB videos, in contrast, are abundant, easy to capture, and represent the most natural form in which human-object interactions are recorded in the wild. They offer a promising path toward scalable HOI reconstruction from everyday videos. Yet monocular reconstruction is fundamentally under-constrained. 

Recent works such as VisTracker~\citep{xie2023visibility}, CARI4D~\citep{cari4d}, and THO~\citep{cari4d} have made important progress in monocular human-object tracking and reconstruction. These methods can recover human and object trajectories in 3D space, and some further introduce contact-aware modules or learned interaction constraints~\citep{wang2025hoitg}. Nevertheless, their outputs often remain visually plausible 4D tracks rather than interaction motions that are physically stable, transferable, or directly usable as humanoid-object animation. We identify three limitations that are especially important for this stricter goal. First, most existing methods are designed around image- or camera-space reconstruction, and are not explicitly organized around a stable, gravity-aware interaction frame, which becomes crucial for moving-camera in-the-wild videos. Second, although contact cues are introduced, they are often coarse, latent, or unreliable, and existing pipelines do not explicitly recover fine hand-object articulation as a first-class part of the interaction. Third, these methods are primarily evaluated as reconstruction systems rather than as teacher trajectories for physical humanoid-object rollout. As a result, even when the recovered human and object trajectories appear reasonable, they may fail to preserve stable contact, functional manipulation, or dynamic plausibility when instantiated in a physical humanoid-object system.

We argue that this formulation misses the central structure of HOI. Human-object interaction is not merely the coexistence of a human trajectory and an object trajectory in 3D space. It is defined by the evolving functional relation between them.We identify this as an \emph{interaction gap} in current monocular HOI reconstruction pipelines. Although they may recover human and object trajectories, they often fail to recover the physical and semantic relation that binds those trajectories into a coherent interaction. Without such articulated interaction, reconstructed HOI animations remain limited for downstream applications that require contact, affordance, and action-level consistency.

To address this gap, we present \textit{HA-HOI} (short for \textbf{H}uman-\textbf{A}nchored \textbf{HOI}), a framework for reconstructing physically plausible 4D HOI animation from in-the-wild monocular videos. Our goal is not merely to recover a visually aligned human-object trajectory, but to produce an interaction sequence that can guide a simulated humanoid toward reasonable object-interaction behavior. Our key insight is that, in most HOI videos, the human is the active agent and the object motion is conditioned on human action. Therefore, the human body should not be treated as just another reconstructed entity in the scene; it should serve as the organizing coordinate system of the interaction. We formulate monocular HOI reconstruction as a \emph{human-first, object-follow} process. 

Specifically, instead of reconstructing the human and object as two independent trajectories in an ambiguous monocular 3D space, we first recover a coherent human motion and use it as the anchor of the interaction. The object is then initialized, constrained, and refined relative to the human motion, where contact, affordance, and interaction can be expressed more naturally. We further incorporate a VLM-based interaction model to propose likely contact surfaces, which allows recovering more meaningful body-object relations while keeping the reconstruction grounded by visual alignment, temporal consistency, and geometric constraints.
Finally, we use the reconstructed sequence as a teacher trajectory for a humanoid controller, requiring the output to not only look plausible in 3D, but also support stable contact and reasonable object interaction under physical rollout.

Empirically, \textit{HA-HOI} reconstructs more coherent and simulation-ready HOI animations from in-the-wild monocular videos. Compared with prior monocular HOI reconstruction methods, our framework improves human-object alignment, contact consistency, temporal stability, and physical rollout quality. Our results suggest that scalable 4D HOI reconstruction should move beyond independent entity tracking and instead treat interaction itself as the organizing principle.

In summary, our contributions are:
\begin{itemize}
    \item We introduce a scalable and flexible framework for transforming in-the-wild monocular HOI videos into simulation-ready humanoid-object interaction trajectories, moving beyond visually aligned 4D tracking toward physically grounded interaction animation, where: 

    \item A human-first, object-follow reconstruction pipeline {\em HA-HOI} is proposed that uses world-grounded human motion as the interaction anchor, reconstructs object motion relative to the human action, and refines fine-grained body-object contact through VLM-guided contact proposals and explicit hand/body optimization.

    \item We demonstrate that \textit{HA-HOI} achieves accurate 4D HOI reconstruction while significantly improves physical plausibility, contact consistency, and temporal stability for humanoid-object simulation.
\end{itemize}

\section{Related Work}

\subsection{3D Foundation Models}

Recent 3D foundation models have greatly advanced monocular reconstruction for humans, objects, and scenes, offering strong priors for body pose, hand geometry, object shape, depth, camera parameters, point maps, tracks, and object pose from in-the-wild images and videos~\citep{goel2023humans, pavlakoshamer, shen2024world, shin2024wham, xia2025reconstructing, zhao2025hunyuan3d, xiang2025structured, xu2024instantmesh, wu2024unique3d, sam3dbody, wu2025motion, wang2025vggt, wang2025moge, wang2024dust3r, wen2024foundationpose, labbe2022megapose, cao2025reconstructing, li2026unish, lu2026track4world}. Yet 4D dynamic scene recovery remains highly challenging. Monocular reconstruction is inherently ambiguous, temporal tracking amplifies geometric inconsistency, and most foundation models are still designed for humans, objects, or scenes in isolation. Consequently, directly composing these priors often produces inconsistent geometry, unstable motion, and weak human-object contact, making temporally and physically consistent 4D HOI reconstruction far from solved.

\subsection{3D HOI Reconstruction}

Static 3D human-object interaction reconstruction has been studied in both hand-object and full-body settings. Existing methods span learning-based joint prediction~\citep{hasson2019learning}, hybrid contact-aware estimation~\citep{yang2021cpf}, and optimization-based reconstruction in the wild~\citep{cao2021reconstructing} for hand-object interaction, as well as spatial fitting~\citep{zhang2020perceiving}], joint neural reconstruction~\citep{xie2022chore}, contact-driven fitting~\citep{cseke2025pico}, foundation-model-guided interaction reasoning~\citep{dwivedi2025interactvlm}, large-model-guided reconstruction~\citep{liu2025easyhoi}], and diffusion-based reconstruction~\citep{li2025scorehoi} for more general human-object interaction. Their development is also supported by increasingly diverse hand-object benchmarks~\citep{shivakumar2019ho, fan2023arctic}. However, many existing methods still rely on supervised training or fitted priors from relatively limited HOI data collected in controlled settings, which restricts generalization to unconstrained in-the-wild scenes. Moreover, although some methods introduce explicit contact losses or contact priors to improve physical plausibility~\citep{hasson2019learning, yang2021cpf, dwivedi2025interactvlm, li2025scorehoi}, they remain largely image-based or frame-wise and do not explicitly enforce temporal consistency, limiting their applicability to stable 4D HOI reconstruction.

\subsection{4D HOI Tracking}

For 4D HOI, recent progress has been enabled by datasets and benchmarks for dynamic human-object capture~\citep{bhatnagar2022behave, huang2024intercap, zhang2024hoi}. Building on these resources, existing methods recover human and object motion from monocular videos through template-based tracking, optimization-based refinement, and transformer-based temporal modeling~\citep{xie2023visibility, zhao2024imhoi, xie2025intertrack, zhang2026end, cari4d}. Despite these advances, explicit interaction modeling remains limited: most methods still rely on coarse geometric cues, and even contact-aware approaches mainly use contact to reduce mesh distance and penetration rather than recover fine-grained contact geometry. Moreover, current 4D HOI methods generally do not incorporate physical simulation, making it difficult to obtain physically grounded interaction dynamics. Although physics-based tracking has been explored for human motion~\citep{tevet2024closd}, it has not been extended to human-object interaction. Motivated by these limitations, we propose a training-free pipeline for more temporally consistent and physically plausible monocular 4D HOI reconstruction.

\section{Methodology}
\label{sec:method}

\begin{figure}
    \vspace{-25pt}
    \centering
    \includegraphics[width=1.02\linewidth]{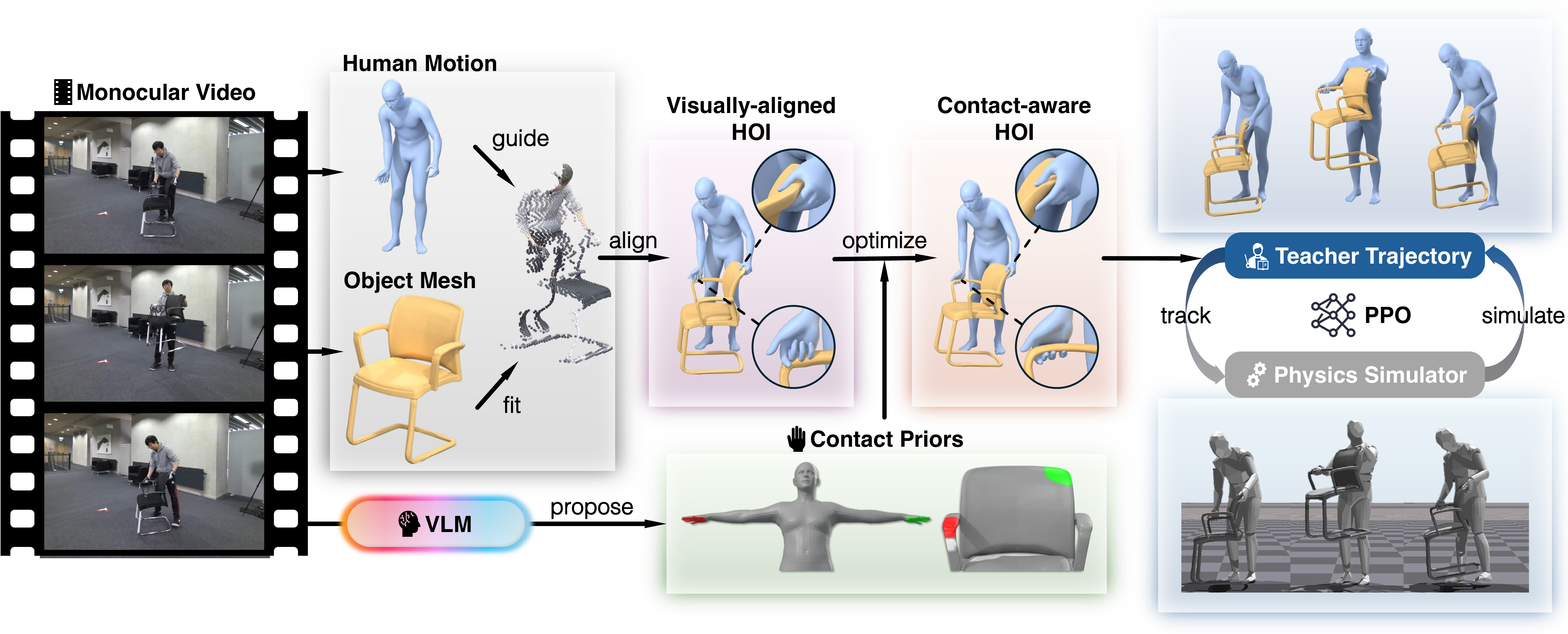}
    \vspace{-20pt}
    \caption{
    \textbf{Overview of \textit{HA-HOI}.}
    From a monocular HOI video, our pipeline recovers human motion and object geometry, aligns them into a visually consistent 4D HOI sequence, and uses VLM-proposed contact priors to refine body-object interaction. The resulting physically plausible HOI trajectory serves as a teacher motion for humanoid-object simulation, enabling stable physical rollout from in-the-wild video.
    }
    \label{fig:pipeline}
    \vspace{-15pt}
\end{figure}

We present \textit{HA-HOI}, a framework that reconstructs physically plausible 4D human-object interaction from a monocular RGB video. 
Given an input video \(\mathcal{V}=\{I_t\}_{t=1}^{T}\), our goal is to recover an interaction trajectory \(\tau=\{H_t,O_t\}_{t=1}^{T}\) that is visually aligned with the video, temporally coherent, contact-consistent, and suitable as a teacher trajectory for physical humanoid-object rollout.

We represent the human state at frame \(t\) as
\(
    H_t=(\boldsymbol{\beta},\boldsymbol{\theta}_t,\boldsymbol{\psi}_t,
    \mathbf{R}^{h}_t,\mathbf{t}^{h}_t),
\)
where \(\boldsymbol{\beta}\) denotes body shape, \(\boldsymbol{\theta}_t\) body pose, \(\boldsymbol{\psi}_t\) hand pose, and \((\mathbf{R}^{h}_t,\mathbf{t}^{h}_t)\) global body orientation and translation. The object is represented by a metric mesh \(\mathcal{M}_o\) and a 6-DoF pose sequence
\(
    O_t=(\mathbf{R}^{o}_t,\mathbf{t}^{o}_t)\in SE(3).
\)

The key design of \textit{HA-HOI} is to reconstruct interaction in a human-centric manner. Instead of independently reconstructing a human track and an object track in an ambiguous monocular 3D space, we first establish the human motion as the interaction anchor. The object is then recovered relative to this anchor, refined using semantic contact proposals, and finally projected into a physical humanoid-object simulation. This yields a pipeline with four stages: human-centric anchoring, object-follow reconstruction, semantic contact refinement, and physics-based interaction projection.

\subsection{Human-Centric Interaction Anchor}
\label{sec:method-human}

Monocular depth and point clouds provide useful local geometric cues, but they are often not stable enough to define the full 4D scene, especially under occlusion and camera motion. We therefore begin by constructing a temporally coherent human motion and use it as the coordinate system of the interaction.

To instantiate this human anchor, we recover world-space body motion represented by SMPL-X~\citep{SMPL-X:2019, shen2024world}. Since hand articulation is critical for human-object interaction, we further estimate MANO hand poses and fuse them into the SMPL-X hand parameters~\citep{MANO:SIGGRAPHASIA:2017, pavlakoshamer}. This gives an articulated human sequence \(\{H_t^0\}_{t=1}^{T}\) that provides both global body motion and fine-grained hand motion.

We then calibrate monocular depth to this human anchor. Let \(D_t\) be the monocular depth prediction~\citep{wang2025moge} at frame \(t\). 
Rather than treating the depth map as an independent global reconstruction, we estimate a scale factor that aligns the unprojected depth points in the human region with the recovered human mesh.
The calibrated depth serves as local geometric evidence for object placement while preserving the temporally coherent scale of the human motion. In this way, the human body defines the metric interaction frame, and depth is used as supporting evidence rather than as the sole scene representation.

\subsection{Object-Follow Reconstruction}
\label{sec:method-object}

Given the human-centric frame, we next recover the manipulated object as an object-follow trajectory. 
The purpose of this stage is not to solve general object reconstruction in isolation, but to obtain an object geometry and pose sequence that remain visually grounded in the video and compatible with the human motion.

We first construct a canonical object mesh \(\mathcal{M}_o^c\) from selected frames in which the object is sufficiently visible. When the object is partially occluded by the human, we use image inpainting to obtain a cleaner object reference before applying a single-image 3D object reconstruction model~\citep{googlenanobanana2, zhao2025hunyuan3d}. 
Since monocular object reconstruction does not reliably determine metric scale, we estimate an object scale \(s_o^\star\) using reference frames with reliable masks and calibrated depth. For each candidate scale \(s\), we align the object to the observed image, render its silhouette and depth, and compare them with the observed object mask and calibrated depth. We choose the scale with the best agreement across reference frames:
\begin{equation}
    s_o^\star =
    \arg\max_s \operatorname{median}_{t \in \mathcal T_{\rm ref}}
    \left[
    \operatorname{IoU}(\hat S_t(s), S_t)
    \exp\left(-\frac{\Delta z_t(s)^2}{2\sigma_d^2}\right)
    \right].
\end{equation}
Here the first term measures silhouette agreement and the second term penalizes depth mismatch after calibration to the human anchor. The median over reference frames makes the estimate robust to occasional occlusion or tracking failure. The metric object mesh is then $\mathcal{M}_o=s_o^\star \mathcal{M}_o^c$.

With the scaled object mesh, we track the object pose across the video using FoundationPose~\citep{wen2024foundationpose}. Since object tracking can fail under occlusion or motion blur, we use mask-depth consistency as a reliability signal. Poses with poor agreement are treated as unreliable and are recovered through interpolation or re-initialization from nearby reliable frames. Finally, the tracked camera-space object poses are transformed into the human-centric world frame.

This stage produces a coarse HOI trajectory
\[
    \tau^0=\{H_t^0,O_t^0\}_{t=1}^{T}.
\]
Although this trajectory is visually grounded and temporally coherent, it may still fail to express the actual interaction: the object may float near the hand, penetrate the body, or touch an incorrect region while still giving the same visual at capture viewpoint. We therefore refine it using semantic contact reasoning.

\subsection{Semantic Contact Refinement}
\label{sec:method-contact}

The coarse trajectory provides human and object motion, but not necessarily a coherent body-object relation. To convert this trajectory into an interaction trajectory, we introduce semantic contact refinement. We use InteractVLM~\citep{dwivedi2025interactvlm} as a contact proposal module. Given representative frames of an interaction segment, it predicts likely contact regions on both the human and the object. These predictions are mapped to vertex sets $\mathcal{C}_h \subset \mathcal{M}_h,
\mathcal{C}_o \subset \mathcal{M}_o,$ where \(\mathcal{M}_h\) is the SMPL-X human mesh and \(\mathcal{M}_o\) is the reconstructed object mesh.

The final interaction is determined by optimizing agreement among contact cues, visual alignment, temporal smoothness, and physical regularity. Starting from \(\tau^0\), we optimize only the variables that directly affect interaction:
\[
    \mathcal{X}
    =
    \{
    \boldsymbol{\theta}^{\mathrm{sel}}_t,
    \boldsymbol{\psi}_t,
    \mathbf{R}^{o}_t,
    \mathbf{t}^{o}_t
    \}_{t=1}^{T},
\]
where \(\boldsymbol{\theta}^{\mathrm{sel}}_t\) denotes selected body joints, \(\boldsymbol{\psi}_t\) the hand pose, and \((\mathbf{R}^{o}_t,\mathbf{t}^{o}_t)\) the object pose. 

We minimize
\begin{equation}
    \min_{\mathcal{X}} \sum_{t=1}^{T} 
    \left( 
    \lambda_c \mathcal{L}_{c} + \lambda_p \mathcal{L}_{p} + \lambda_s \mathcal{L}_{s} + \lambda_a \mathcal{L}_{a} + \lambda_r \mathcal{L}_{r} 
    \right),
\end{equation}

where \(\mathcal{L}_{c}\) encourages contact consistency, \(\mathcal{L}_{p}\) penalizes penetration, \(\mathcal{L}_{s}\) encourages temporal smoothness, \(\mathcal{L}_{a}\) anchors the refinement to the coarse reconstruction, and \(\mathcal{L}_{r}\) regularizes body and hand poses.

The core term is the contact loss. During early iterations, we use a bidirectional nearest-neighbor distance between predicted human and object contact regions to obtain stable coarse contact. Once the interaction is approximately aligned, we switch to an object signed-distance-field contact objective:
\begin{equation}
    \mathcal{L}_{c}=
    \sum_{s\in\{l,r\}}
    \frac{1}{|\mathcal{C}^{s}_{h}|}
    \sum_{i\in\mathcal{C}^{s}_{h}}\Phi_o\left((\mathbf{R}^{o}_t)^\top(\mathbf{x}^{h}_{t,i}-\mathbf{t}^{o}_t)\right)^2, 
\end{equation}
where \(\mathbf{x}^{h}_{t,i}\) is a human contact vertex and \(\Phi_o\) is the signed distance field of the object in its canonical coordinate frame. This objective pulls predicted human contact vertices onto the object surface while allowing the object and local hand/body pose to adapt.

Our progressive otimization proceeds as follows:  First, we adjust object depth under the fixed human motion, which corrects the dominant monocular contact error. Next, we refine the full object pose while preserving the global human anchor. We then jointly optimize object pose, hand pose, and selected body joints so that the object and body can adapt to each other. Finally, with the object fixed, we refine wrists and fingers using contact, coverage, penetration, non-contact repulsion, and hand-pose priors.

The result is a contact-refined kinematic trajectory
\[
    \tau^\star=\{H_t^\star,O_t^\star\}_{t=1}^{T},
\]
which is visually aligned and geometrically interaction-aware. However, it is still a kinematic reconstruction. A trajectory can satisfy geometric contact losses while remaining unstable under gravity or rigid-body dynamics. We therefore use physical simulation as a final interaction projection step.

\subsection{Physics-Based Interaction Projection}
\label{sec:method-physics}
We use physics-based tracking as the final projection step by instantiating a corresponding humanoid-object system in a physics simulator. We convert the optimized human sequence \(H_t^\star\) into a humanoid reference motion with generalized coordinates \(\bar{\mathbf{q}}_t\), and use the optimized object sequence \(O_t^\star\) as the reference object trajectory \(\bar{O}_t\). The object is modeled as a dynamic rigid body, and the humanoid is controlled through joint-level actuation.

Rather than directly replaying the kinematic reconstruction, we train a sequence-specific residual controller. At simulation step \(t\), the policy observes the current simulated state and the reference state in a root-relative coordinate frame. It outputs a residual action \(\mathbf{a}_t\), which modifies the reference joint target:
\[
    \mathbf{q}^{\mathrm{tar}}_t=\bar{\mathbf{q}}_t+\alpha \mathbf{a}_t,
\]
where \(\alpha\) controls the maximum deviation from the reconstructed reference. The humanoid is driven toward \(\mathbf{q}^{\mathrm{tar}}_t\) using PD control, while the object evolves under contact and gravity.

This formulation treats the reconstructed trajectory as a strong teacher while allowing the simulator to correct small errors that would otherwise cause slipping, jitter, or unstable contact. The tracking reward balances motion fidelity and interaction stability:
\begin{equation}
    r_t=w_h r^h_t+w_o r^o_t+w_{\mathrm{hand}} r^{\mathrm{hand}}_t+w_{\mathrm{contact}} r^{\mathrm{contact}}_t+w_{\mathrm{vel}} r^{\mathrm{vel}}_t .
\end{equation}
The human tracking term encourages the simulated humanoid to follow the reconstructed body motion. The object term rewards agreement with the reference object trajectory. The hand term emphasizes grasp-related articulation. The contact term encourages predicted body-object contacts to remain active during rollout, and the velocity term discourages implausible accelerations.

After training, we roll out the controller once and convert the simulated states back into SMPL-X parameters and object poses. The final output is
\[
\tau^{\mathrm{phys}}
=
\{H_t^{\mathrm{phys}},O_t^{\mathrm{phys}}\}_{t=1}^{T},
\]
a physically plausible humanoid-object animation that remains close to the input video while supporting stable interaction under simulation.

\section{Experiments}

\subsection{Experiments Setup}

\begin{table}[t]
\centering
\vspace{-15pt}
\setlength{\tabcolsep}{4.0pt}
\renewcommand{\arraystretch}{1.05}
\resizebox{0.86\textwidth}{!}{
\begin{tabular}{lcccccccc}
\toprule
\textbf{Method} 
& Moving Cam. 
& Template-free
& CD-h$\downarrow$ 
& CD-o$\downarrow$ 
& CD-c$\downarrow$ 
& Acc-h$\downarrow$ 
& Acc-o$\downarrow$ 
& Pen.$\downarrow$ \\
\midrule
CHORE          & \xmark & \xmark & 25.82 & 28.94 & 26.52 & 2.75 & 4.59 & --    \\
InterTrack     & \xmark & \cmark & 25.71 & 47.66 & 30.20 & 5.23 & 5.64 & --    \\
HOI-TG         & \xmark & \xmark & 21.19 & 24.70 & 22.67 & 2.44 & 3.76 & --    \\
VisTracker     & \xmark & \xmark & 11.24 & 12.58 & 11.63 & 0.54 & 0.77 & 0.079    \\
CARI4D$^\ast$  & \xmark & \cmark & 7.51  & \textbf{11.93} & \textbf{9.06} & 1.07 & \textbf{0.41} & 0.084 \\
THO            & \xmark & \xmark & 8.49  & 13.36 & 10.06 & 0.79 & 1.02 & -- \\
\midrule
\textbf{Ours}  & \cmark & \cmark & \textbf{7.11} & 15.68 & 10.74 & \textbf{0.52} & 0.59 & \textbf{0.013} \\
\bottomrule
\end{tabular}
}
\vspace{2pt}
\caption{
Quantitative comparison on BEHAVE. We report Chamfer distance (in cm) for human, object, and combined human-object surfaces, acceleration error (in cm$/$s$^2$) for temporal consistency, and human-object penetration (in cm). 
Unlike template-based tracking methods, our method reconstructs the object geometry and does not require a ground-truth object template at test time. 
$^\ast$CARI4D is evaluated on 62 selected sequences.
}
\label{tab:quantitative_comparison}
\vspace{-15pt}
\end{table}

\paragraph{Datasets.}
We evaluate our method on the BEHAVE dataset~\citep{bhatnagar2022behave}, a full-body human-object interaction benchmark captured with multi-view RGB-D sensors and released with pseudo-ground-truth SMPL body and object registrations. 
Following the standard monocular HOI reconstruction protocol used by prior work~\citep{xie2023visibility,cari4d,zhang2026end}, we use the extended BEHAVE annotations, where human and object registrations are provided at 30 FPS. 
We adopt the official split with 217 training sequences and 82 test sequences, and report results on the test set. Although BEHAVE is captured with multi-view RGB-D sensors, our method takes monocular RGB video as input.

\paragraph{Metrics and protocol.}

We evaluate 4D reconstruction quality using Chamfer distance between reconstructed and ground-truth surfaces. CD-h, CD-o, and CD-c denote Chamfer distance for the human, object, and combined human-object meshes, respectively. 

We further report acceleration error for temporal consistency, where Acc-h, Acc-o denote the mean per-joint human and object translation error. 
To measure interaction plausibility, we report human-object penetration as the average negative signed distance from human-body vertices to the posed object surface. Lower penetration indicates fewer physically implausible human-object intersections.

All results are produced by running the full pipeline end-to-end from a monocular RGB video. We do not use BEHAVE-specific training, ground-truth object templates, object pose initialization, contact annotations, or sequence-level manual adjustment. 
Unless otherwise specified, all reconstruction metrics are computed before physics simulation. The physics stage introduces controller dynamics and contact responses that may intentionally modify the kinematic trajectory; we therefore evaluate simulation behavior separately using rollout-based criteria.
Following recent 4D HOI evaluation protocols~\citep{xie2023visibility,cari4d}, we align the reconstructed human mesh in the first frame to the ground-truth human mesh and apply the same transformation to the entire sequence. This avoids per-frame alignment that could hide temporal drift, while also avoiding alignment through the object mesh whose topology may differ from the ground truth. We convert the final human reconstruction from SMPL-X to SMPL-H before evaluation for compatibility with the ground-truth annotations and prior baselines.

\subsection{Quantitative Comparison on BEHAVE}

Table~\ref{tab:quantitative_comparison} compares \textit{HA-HOI} with recent monocular HOI reconstruction and tracking methods, including CHORE~\citep{xie2022chore}, InterTrack~\citep{xie2025intertrack}, HOI-TG~\citep{wang2025hoitg}, VisTracker~\citep{xie2023visibility}, CARI4D~\citep{cari4d}, and THO~\citep{zhang2026end}. 
Different from many prior methods that rely on a ground-truth object template at test time or learn BEHAVE-specific interaction priors, our method runs the full pipeline end-to-end from monocular RGB video. 
It does not assume a predefined object mesh, object pose initialization, contact annotation, or sequence-level manual adjustment.

Despite this more general setting, \textit{HA-HOI} achieves competitive reconstruction accuracy across all metrics. 
In particular, our method obtains the best human reconstruction accuracy, reducing CD-h from 7.51 to 7.11, and also achieves the lowest human acceleration error, reducing Acc-h from 0.54 to 0.52. 
These gains support our human-centered formulation: by first recovering a coherent human motion and using it as the anchor of the interaction, the reconstruction remains temporally stable while preserving accurate body geometry.

More importantly, \textit{HA-HOI} substantially improves the interaction-specific physical plausibility of the reconstructed sequence. 
Our method achieves the lowest penetration score, reducing human-object penetration from 0.084 to 0.013 compared with the strongest reported baseline. 
This result is central to our objective: the reconstructed sequence should not merely align human and object surfaces in visual space, but should preserve physically meaningful contact relations that can serve as a teacher trajectory for downstream humanoid-object simulation. 
Although template-based methods can achieve lower object Chamfer distance by leveraging stronger object geometry priors, our results show that optimizing for interaction consistency leads to more contact-faithful and simulation-ready 4D HOI reconstruction.

\subsection{Qualitative Results}

\begin{figure}[t]
    \vspace{-20pt}
    \centering
    \includegraphics[width=\textwidth]{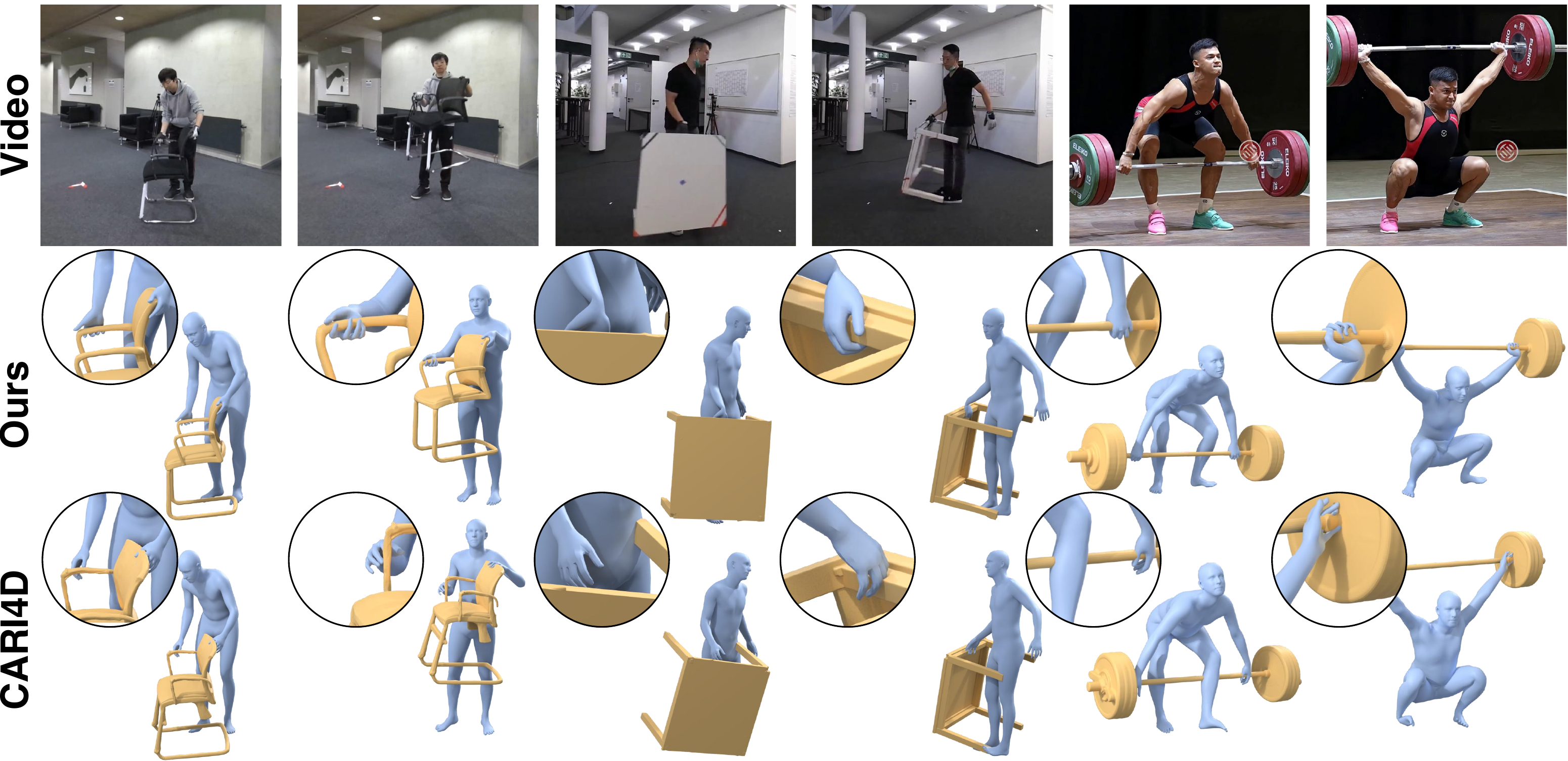}
    \caption{Qualitative comparison with CARI4D on BEHAVE test sequences, BEHAVE training sequences, and in-the-wild videos. 
    CARI4D often fails to maintain functional hand-object relations, leading to floating, missing, or ambiguous contacts. 
    In contrast, \textit{HA-HOI} reconstructs contact-aware HOI animations with articulated hands, stable object support, and more physically plausible interaction geometry. 
    }
    \label{fig:quali}
    \vspace{-15pt}
\end{figure}

Figure~\ref{fig:quali} provides qualitative comparisons on three representative videos: a BEHAVE test sequence, a BEHAVE training sequence, and an in-the-wild interaction video. 
Our results are shown after physics simulation, which makes the comparison more demanding than visual reconstruction alone.

\textit{HA-HOI} produces stable, temporally coherent, and visually aligned HOI animations across all three settings. 
More importantly, it preserves the functional relation between the human body and the object. 
In the chair and table examples, our method maintains plausible hand-object contact and object placement throughout the interaction. 
In the weightlifting example, our reconstruction preserves the grasp-like relation between both hands and the barbell, allowing the object to remain physically tied to the human action after simulation.

Compared with CARI4D, the advantage of \textit{HA-HOI} is most visible around contact regions. 
CARI4D often reconstructs a visually-aligned global human-object layout from camera view, but the body parts do not reliably explain the interaction: the object may be close to the body while the fingers float, penetrate, or fail to establish stable contact. 
This is a critical failure mode for previous 4D HOI reconstruction, because the interaction is not defined only by where the human and object are, but by how the human manipulates the object.

These qualitative results complement the quantitative findings in Table~\ref{tab:quantitative_comparison}. 
Standard Chamfer and acceleration metrics capture coarse geometry and temporal smoothness, but they do not fully reflect hand articulation, grasp formation, or whether the reconstructed interaction can survive physical rollout. 
By explicitly anchoring object reconstruction to human motion and refining contact-aware geometry, \textit{HA-HOI} recovers HOI animations that are not only visually plausible, but also physically meaningful after simulation. 
This supports our central claim that monocular 4D HOI reconstruction should move beyond independent human-object tracking and recover the interaction itself.

\subsection{Ablation Study}

\begin{wraptable}{r}{0.50\textwidth}
\vspace{-8pt}
\centering
\scriptsize
\setlength{\tabcolsep}{2.2pt}
\renewcommand{\arraystretch}{1.05}
\resizebox{\linewidth}{!}{
\begin{tabular}{lcccccc}
\toprule
~ 
& CD-h$\downarrow$ 
& CD-o$\downarrow$ 
& CD-c$\downarrow$ 
& Acc-h$\downarrow$ 
& Acc-o$\downarrow$
& Pen.$\downarrow$ \\
\midrule
\textbf{Ours}       & 7.11  & 15.68 & 10.74 & 0.52 & 0.59 & 0.013 \\
w/ GT mesh          & 7.13  & 13.96 & 10.12 & 0.52 & 0.58 & 0.007 \\
w/ GT cam.          & 7.08  & 15.19 & 10.45 & 0.52 & 0.57 & 0.014 \\
w/o contact opt.    & 7.10  & 15.49 & 10.74 & 0.51 & 0.49 & 0.097 \\
\bottomrule
\end{tabular}
}
\vspace{-4pt}
\caption{
Ablation study on BEHAVE using ground truth meshes and camera parameters, and removing the contact optimization. 
}
\label{tab:ablation}
\end{wraptable}
We conduct ablation studies to isolate the effect of object geometry, camera estimation, and contact refinement. 
Using the ground-truth object mesh improves CD-o and CD-c, showing that part of the remaining object error comes from open-world object reconstruction rather than interaction optimization. 
However, the improvement is moderate, and the human metrics remain almost unchanged, indicating that the proposed human-centered reconstruction is not dependent on object templates.

Replacing the estimated camera with the ground-truth camera also brings only a small improvement in object and combined Chamfer distance. 
This suggests that camera estimation is not the dominant bottleneck in our pipeline. 
More importantly, \textit{HA-HOI} remains stable under the fully automatic setting, where neither the ground-truth object mesh nor the ground-truth camera is provided.

Removing contact refinement can even marginally improve some visual-alignment scores. 
This is precisely the failure mode we aim to address: surface-distance metrics can favor geometrically close but physically invalid reconstructions. 
Without contact refinement, penetration increases from 0.013 to 0.097, indicating that the human and object are better aligned as independent surfaces but worse aligned as an interaction. 
Our contact refinement therefore introduces a deliberate bias toward physical compatibility, sacrificing no meaningful reconstruction accuracy while substantially reducing implausible intersections. 
This supports our central claim that monocular 4D HOI should be evaluated and optimized as an interaction, not merely as two visually aligned trajectories.

\section{Discussions}

\subsection{Limitations and Future Work}
\textit{HA-HOI} demonstrates that monocular HOI videos can be transformed into simulation-ready interaction trajectories. However, the reconstruction stage remains largely vision-based and severe occlusion may lead to inaccurate human body pose, or object geometry. 
A natural next step is to deploy the reconstructed interactions by \textit{HA-HOI} on real humanoid robots. Our current physics-based rollout provides an intermediate testbed for simulation-ready HOI, but real-world execution introduces additional challenges, including sim-to-real transfer, hardware limits, sensing noise, and contact robustness. Bridging this gap would further advance the central vision of this work: using general monocular HOI videos as scalable demonstrations for learning humanoid-object behavior.

\subsection{Conclusion}
We introduced \textit{HA-HOI}, a framework that converts in-the-wild monocular HOI videos into physically plausible 4D human-object interaction trajectories. Our central claim is that HOI reconstruction should not stop at recovering visually coherent human and object tracks. Instead, it should recover the interaction relation that makes those tracks meaningful and usable. To this end, \textit{HA-HOI} adopts a human-first, object-follow formulation, uses VLM-based contact proposals to refine body-object relations, and projects the resulting trajectory into physics-based humanoid-object simulation.

Through reconstruction and simulation experiments, we show that this formulation improves human-object alignment, contact consistency, temporal stability, and physical rollout quality. More broadly, our work points to a new direction for monocular 4D HOI reconstruction: using ordinary videos not only as visual observations, but as scalable sources of interaction demonstrations. We believe this shift from trajectory reconstruction to physically grounded interaction animation is an important step toward learning humanoid-object behaviors from the open visual world.


\newpage
{
    \small
    \bibliographystyle{ieeenat_fullname}
    \bibliography{main}
}


\newpage
\appendix
\addcontentsline{toc}{section}{Appendix Index}

This appendix provides additional qualitative results, implementation details, and evaluation protocols that support the main paper.

\begin{itemize}

    \item \textbf{Appendix~\ref{sec:technical_appendices}: Technical Appendices.}
    Implementation details of the reconstruction and simulation pipeline, including:
    \begin{itemize}
        \item \textbf{Appendix~\ref{sec:object_scale_estimator}: Object Scale Estimator.}
        Monocular object-scale estimation using silhouette IoU and depth consistency.

        \item \textbf{Appendix~\ref{sec:object_tracker}: Object Tracker.}
        FoundationPose tracking with health-gated recovery and re-registration.

        \item \textbf{Appendix~\ref{sec:contact_optimization}: Contact Optimization.}
        Four-stage human-object contact refinement and loss definitions.

        \item \textbf{Appendix~\ref{sec:convex_decomposition}: Convex Decomposition.}
        Collision-geometry preparation for physics simulation.

        \item \textbf{Appendix~\ref{sec:physics_simulation}: Physics Simulation.}
        State/action space, reward design, and PPO training details.
    \end{itemize}

    \item \textbf{Appendix~\ref{sec:evaluation_details}: Evaluation Details.}
    Metric definitions, repeated-run protocol, and computational requirements.
\end{itemize}

\section{Technical Appendices}
\label{sec:technical_appendices}

\subsection{Object Scale Estimator}
\label{sec:object_scale_estimator}

Monocular depth estimation produces metric-ambiguous depth maps. To resolve the scale ambiguity for object reconstruction, we employ a unified scale estimator that combines silhouette IoU and depth consistency:

\begin{equation}
s^* = \operatorname{argmax}_{s \in [s_{\min}, s_{\max}]} \text{IoU}(M_{\text{pred}}(s), M_{\text{obs}}) \cdot \exp\left(-\frac{1}{2}\left(\frac{\Delta z(s)}{\sigma_{\text{depth}}}\right)^2\right)
\end{equation}

where $s$ is the candidate scale factor, $M_{\text{pred}}(s)$ is the rendered object silhouette at scale $s$, $M_{\text{obs}}$ is the observed mask from SAM3, $\Delta z(s)$ is the mean depth residual between rendered and observed depth, and $\sigma_{\text{depth}}=0.05$m is the depth tolerance.

We search over a discrete set of scales (default: 13 steps from 0.5 to 1.5) on a subset of clean reference frames (default: 8 frames) selected by quality gates. A self-consistency re-pass is triggered if $|s^* - 1| > 0.25$, re-running FoundationPose registration with the scaled mesh.

\subsection{Object Tracker}
\label{sec:object_tracker}

\subsubsection{FoundationPose + Healthy Gated Tracker}

We extend FoundationPose~\cite{wen2024foundationpose} with a health-gated tracking mechanism to improve robustness. The tracker maintains a health score $h_t \in [0,1]$ for each frame based on:

\begin{equation}
h_t = \alpha_{\text{depth}} \cdot h_t^{\text{depth}} + \alpha_{\text{mask}} \cdot h_t^{\text{mask}}
\end{equation}

where $h_t^{\text{depth}}$ measures depth consistency and $h_t^{\text{mask}}$ measures mask IoU between rendered and observed masks.

The tracker operates in three modes:
\begin{itemize}
    \item \textbf{Healthy tracking} ($h_t > \theta_{\text{ok}}=0.20$): Standard FoundationPose tracking
    \item \textbf{Degraded tracking} ($\theta_{\text{deg}}=0.15 < h_t \leq \theta_{\text{ok}}$): Increase refinement iterations
    \item \textbf{Re-registration} ($h_t \leq \theta_{\text{deg}}$): Trigger full pose estimation from scratch
\end{itemize}

A cooldown period (default: 10 frames) prevents rapid re-registration oscillations. Optionally, frozen tracking stretches (low-confidence segments) can be interpolated from neighboring healthy frames.

\subsection{Contact Optimization}
\label{sec:contact_optimization}

\subsubsection{Four-Stage Optimization}

Our contact optimization proceeds in four stages to progressively refine the reconstruction:

\paragraph{Stage 1: Depth Pre-optimization} Optimize SMPL translation to align with depth maps while maintaining contact constraints:
\begin{equation}
\mathcal{L}_{\text{depth-pre}} = w_c \mathcal{L}_{\text{contact}} + w_p \mathcal{L}_{\text{pen}} + w_t \mathcal{L}_{\text{temporal}}
\end{equation}
Iterations: 100, LR: 0.005, weights: $w_c=50$, $w_p=20$, $w_t=3$.

\paragraph{Stage 2: Object Pose Pre-optimization} Optimize object 6-DoF poses with contact and temporal constraints:
\begin{equation}
\mathcal{L}_{\text{obj-pre}} = w_c \mathcal{L}_{\text{contact}} + w_p \mathcal{L}_{\text{pen}} + w_{t,\text{trans}} \mathcal{L}_{\text{temporal}}^{\text{trans}} + w_{t,\text{rot}} \mathcal{L}_{\text{temporal}}^{\text{rot}} + w_a \mathcal{L}_{\text{anchor}}
\end{equation}
Iterations: 150, LR: $\text{trans}=0.003$, $\text{rot}=0.001$, weights: $w_c=50$, $w_p=20$, $w_{t,\text{trans}}=3$, $w_{t,\text{rot}}=2$, $w_a=1$.

\paragraph{Stage 3: Joint Optimization} Jointly optimize SMPL body pose and object poses:
\begin{equation}
\mathcal{L}_{\text{joint}} = w_c \mathcal{L}_{\text{contact}} + w_p \mathcal{L}_{\text{pen}} + w_{t,\text{smpl}} \mathcal{L}_{\text{temporal}}^{\text{smpl}} + w_{t,\text{obj}} \mathcal{L}_{\text{temporal}}^{\text{obj}} + w_a \mathcal{L}_{\text{anchor}}
\end{equation}
Iterations: 300, LR: $\text{body}=0.0005$, $\text{obj-trans}=0.003$, $\text{obj-rot}=0.001$.

\paragraph{Stage 4: Hand Refinement} Refine hand poses using HaMeR priors:
\begin{equation}
\mathcal{L}_{\text{hand}} = w_c \mathcal{L}_{\text{contact}} + w_p \mathcal{L}_{\text{hand-pen}} + w_{\text{pca}} \mathcal{L}_{\text{hand-pca}} + w_t \mathcal{L}_{\text{temporal}}^{\text{hand}}
\end{equation}
Iterations: 200, LR: 0.01, weights: $w_{\text{pca}}=3$, hand PCA components: 12.

\subsubsection{Loss Function Details}

\paragraph{Contact Loss} Bidirectional chamfer distance between contact regions:
\begin{equation}
\mathcal{L}_{\text{contact}} = \frac{1}{|V_h|} \sum_{v \in V_h} \min_{u \in V_o} \|\mathbf{v} - \mathbf{u}\|_2^2 + \frac{1}{|V_o|} \sum_{u \in V_o} \min_{v \in V_h} \|\mathbf{u} - \mathbf{v}\|_2^2
\end{equation}
where $V_h$ and $V_o$ are vertices in human and object contact regions.

\paragraph{Penetration Loss} Penalize interpenetration using signed distance:
\begin{equation}
\mathcal{L}_{\text{pen}} = \sum_{v \in V_{\text{pen}}} \max(0, -d(v))^2
\end{equation}
where $d(v) < 0$ indicates penetration depth.

\paragraph{Temporal Smoothness} Acceleration penalty:
\begin{equation}
\mathcal{L}_{\text{temporal}} = \sum_{t=1}^{T-1} \|\theta_t - 2\theta_{t-1} + \theta_{t-2}\|_2^2
\end{equation}

\subsection{Convex Decomposition}
\label{sec:convex_decomposition}

For physics simulation in IsaacGym, both human and object meshes require convex decomposition:

\begin{itemize}
    \item \textbf{Object}: V-HACD with parameters: max convex hulls: 32, max vertices per hull: 64, resolution: 300000
    \item \textbf{Humanoid}: Custom MuJoCo XML articulated body generated from SMPL-X with convex collision geometry
\end{itemize}

The decomposed collision geometry is used for contact detection and physics simulation, while the original high-resolution mesh is used for rendering.

\subsection{Physics Simulation}
\label{sec:physics_simulation}

Our physics simulator uses IsaacGym with PPO reinforcement learning. The humanoid is represented as a custom MuJoCo XML articulated body generated from SMPL-X with $\beta$-dependent link dimensions and masses.

\subsubsection{State and Action Space}

\paragraph{State} $s_t \in \mathbb{R}^{169}$:
\begin{itemize}
    \item Root position and orientation (7D: pos + quat)
    \item Joint positions and velocities (63D $\times$ 2)
    \item Object position, orientation, linear and angular velocities (13D)
    \item Phase variable (1D)
\end{itemize}

\paragraph{Action} $a_t \in \mathbb{R}^{63}$: Target joint positions (PD control).

\subsubsection{Reward Function}

\begin{equation}
r_t = w_{\text{pose}} r_{\text{pose}} + w_{\text{vel}} r_{\text{vel}} + w_{\text{contact}} r_{\text{contact}} + w_{\text{obj}} r_{\text{obj}}
\end{equation}

where:
\begin{itemize}
    \item $r_{\text{pose}} = \exp(-\|\mathbf{q}_t - \mathbf{q}_t^{\text{ref}}\|_2^2)$: Pose tracking reward
    \item $r_{\text{vel}} = \exp(-\|\dot{\mathbf{q}}_t - \dot{\mathbf{q}}_t^{\text{ref}}\|_2^2)$: Velocity tracking reward
    \item $r_{\text{contact}} = \mathbbm{1}[\text{contact matches reference}]$: Contact reward (warmup: 3 steps)
    \item $r_{\text{obj}} = \exp(-\|\mathbf{p}_t^{\text{obj}} - \mathbf{p}_t^{\text{obj,ref}}\|_2^2)$: Object tracking reward
\end{itemize}

Weights: $w_{\text{pose}}=0.5$, $w_{\text{vel}}=0.1$, $w_{\text{contact}}=0.2$, $w_{\text{obj}}=0.2$.

\subsubsection{Training Details}

\begin{itemize}
    \item \textbf{Algorithm}: PPO with GAE ($\lambda=0.95$, $\gamma=0.95$)
    \item \textbf{Network}: Actor-critic with separate networks
    \begin{itemize}
        \item Actor: 5-layer MLP (2048-2048-1024-1024-512 hidden units, GELU activation)
        \item Critic: 5-layer MLP (2048-2048-1024-1024-512 hidden units, GELU activation)
    \end{itemize}
    \item \textbf{Optimization}: Adam optimizer, actor LR: $5 \times 10^{-6}$, critic LR: $1 \times 10^{-4}$
    \item \textbf{Training schedule}: Horizon: 8, batch size: 256, optimization epochs: 5
    \item \textbf{Parallel envs}: 512 (configurable)
    \item \textbf{Episode length}: 150 frames (configurable)
    \item \textbf{Max training epochs}: 1,000,000 (varies by sequence complexity)
\end{itemize}

\section{Evaluation Details}
\label{sec:evaluation_details}

\subsection{Evaluation Metrics Definitions}

\subsubsection{Chamfer Distance}

For meshes $\mathcal{M}_{\text{pred}}$ and $\mathcal{M}_{\text{gt}}$ with vertex sets $V_{\text{pred}}$ and $V_{\text{gt}}$:

\begin{equation}
\text{CD}(\mathcal{M}_{\text{pred}}, \mathcal{M}_{\text{gt}}) = \frac{1}{|V_{\text{pred}}|} \sum_{v \in V_{\text{pred}}} \min_{u \in V_{\text{gt}}} \|v - u\|_2 + \frac{1}{|V_{\text{gt}}|} \sum_{u \in V_{\text{gt}}} \min_{v \in V_{\text{pred}}} \|u - v\|_2
\end{equation}

\subsubsection{Acceleration Error}

For joint positions $\mathbf{j}_t$ at frame $t$:

\begin{equation}
\text{Acc} = \frac{1}{T-2} \sum_{t=1}^{T-2} \left\| (\mathbf{j}_{t+1} - 2\mathbf{j}_t + \mathbf{j}_{t-1}) - (\mathbf{j}_{t+1}^{\text{gt}} - 2\mathbf{j}_t^{\text{gt}} + \mathbf{j}_{t-1}^{\text{gt}}) \right\|_2
\end{equation}

\subsubsection{Penetration}

In our setting, we measure human-body vertices penetrating the object:

\begin{equation}
\mathrm{Pen} =
\frac{1}{T |V_{\mathrm{human}}|}
\sum_{t=1}^{T}
\sum_{v \in V_{\mathrm{human}}^{t}}
\max\bigl(0, -d_{\mathrm{obj}}^{t}(v)\bigr),
\end{equation}
where \(V_{\mathrm{human}}^{t}\) are the human mesh vertices at frame (t), and \(d_{\mathrm{obj}}^{t}(v)\) is the signed distance from human vertex (v) to the posed object mesh at frame (t). Negative values indicate that the human vertex lies inside the object. The result is reported in centimeters.

\subsection{Dataset Statistics}
To improve the reliability of our empirical evaluation, we repeat each experiment over 10 independent runs and report the averaged results in the main tables. The repeated runs use the same input videos, dataset split, evaluation protocol, and metric definitions, while accounting for stochasticity introduced by optimization, initialization, tracking recovery, and sequence-specific physics simulation. 

This repeated-run protocol is used for the proposed method and its ablation variants. Since several baseline results are taken from prior published evaluations under the standard BEHAVE protocol, direct paired statistical tests against all baselines are not always possible. We therefore use the 10-run averages to assess the stability of our own pipeline and ablations, and report aggregate metrics following the evaluation convention of prior monocular 4D HOI reconstruction work. We also report the $\pm\ 1$SD error bar below.

\begin{table}[h]
\centering
\setlength{\tabcolsep}{4.0pt}
\renewcommand{\arraystretch}{1.05}
\resizebox{0.86\textwidth}{!}{
\begin{tabular}{lcccccc}
\toprule
\textbf{Method} 
& CD-h 
& CD-o 
& CD-c 
& Acc-h 
& Acc-o 
& Pen. \\
\midrule
\textbf{Ours} & 7.11$\pm$0.01 & 15.68$\pm$0.11 & 10.74$\pm$0.09 & 0.52$\pm$0.01 & 0.59$\pm$0.02 & 0.013$\pm$0.001 \\
\bottomrule
\end{tabular}
}
\vspace{2pt}
\caption{
Quantitative experiments on BEHAVE test set with $\pm\ 1$SD confidence interval.
}
\label{tab:error}
\vspace{-15pt}
\end{table}

\subsection{Computational Requirements}
We measure the computational cost of the full framework on a single NVIDIA H200 GPU workstation with 1.8TB system RAM. On BEHAVE, processing a 1500-frame monocular RGB video takes approximately 47 minutes from input video to the final simulation-ready human-object interaction trajectory. 

The average runtime is therefore approximately 1.88 seconds per frame for a 1500-frame sequence. Actual runtime may vary depending on sequence length. In particular, heavily occluded sequences may require additional tracking recovery or contact refinement, while shorter and visually cleaner sequences are processed more efficiently.

\end{document}